\documentclass[journal]{IEEEtran}

\ifCLASSINFOpdf
\else
\fi

\usepackage{amsthm}
\usepackage{amsmath}
\usepackage{amssymb}
\usepackage{amsbsy}

\newtheorem{theorem}{Theorem}

\usepackage{algpseudocode}
\usepackage{algorithm}
\usepackage{algorithmicx}
\usepackage{balance}
\usepackage{graphicx}
\usepackage{subfig}

\newcommand*{\MYI}{\skew{3}{\hat}{i}}
\begin{document}
%
\title{Particle Swarm Optimization: Stability Analysis using $N$-Informers under Arbitrary Coefficient Distributions} 
%
%
%

\author{Christopher~W~Cleghorn,~Belinda~Stapelberg
\thanks{C.W. Cleghorn is with the Department of Computer Science, University of Pretoria, South Africa, Email: ccleghorn@cs.up.ac.za}
\thanks{B. Stapelberg is with the Department of Mathematics and Applied Mathematics, University of Pretoria, South Africa, {Email: belinda.stapelberg@up.ac.za}}
\thanks{Manuscript received March 31, 2020}}

\markboth{}%
{Shell \MakeLowercase{\textit{et al.}}: Bare Demo of IEEEtran.cls for IEEE Journals}
%



\providecommand{\norm}[1]{\lVert#1\rVert}
\maketitle

\begin{abstract}
This paper derives, under minimal modelling assumptions, a simple to use theorem for obtaining both order-$1$ and order-$2$ stability criteria for a common class of particle swarm optimization (PSO) variants. Specifically, PSO variants that can be rewritten as a finite sum of stochastically weighted difference vectors between a particle's position and swarm informers are covered by the theorem.  Additionally, the use of the derived theorem allows a PSO practitioner to obtain stability criteria that contains no artificial restriction on the relationship between control coefficients. Almost all previous PSO stability results have provided stability criteria under the restriction that the social and cognitive control coefficients are equal; such restrictions are not present when using the derived theorem. Using the derived theorem, as demonstration of its ease of use, stability criteria are derived without the imposed restriction on the relation between the control coefficients for three popular PSO variants.
\end{abstract}


%
\IEEEpeerreviewmaketitle

\section{Introduction}
%
%
%
%
\IEEEPARstart{T}{he} particle swarm optimization (PSO) algorithm, originally developed by Kennedy and Eberhart \cite{OriginalPSOPaper}, has become a widely used optimization technique \cite{PoliAppReview}. Given PSO's popularity, it has undergone a considerable amount of theoretical investigation, to list just a few, \cite{FirstTheoryPaper,KennedyContriction,Jiang,Fernadez_CPSO_Arb_Dis,ChrisClegh,weakStagnation,ChrisCleghPSOGeneralStabilityTheorem}. 

There are a number of aspects of PSO behaviour that can be investigated from a theoretical perspective. However, the focus of this paper is on the criteria needed for order-$1$ and order-$2$ stability of PSO particles. Specifically, order-$1$ and order-$2$ stability occurs when particle positions converge to a constant in first and second order moment respectively \cite{CleghWhatIsStability}\footnote{Some authors have considered the stricter condition where the second order moment converges to zero, a detailed justification for using convergence to a constant second order moment is provided in \cite{CleghWhatIsStability}.} . The vast majority of theoretical studies have focused on reducing the modelling assumption used to obtain the stability criteria for PSO with inertia (referred to as canonical PSO (CPSO) in this paper), as proposed by Shi and Eberhart \cite{InertiaIntroduction}. A detailed discussion of the systematic weakening of these modelling assumption can be found in \cite{ChrisCleghPSOGeneralStabilityTheorem}.	

The focus of this paper is instead, on  providing an easy to use theorem for obtaining stability criteria for PSO variants, while using the minimal modelling assumptions proposed by  Cleghorn and Engelbrecht \cite{ChrisCleghPSOGeneralStabilityTheorem}. As such, the general aim of this paper is to provide a theorem that allows a researcher to still obtain stability criteria even if they have made alterations, within reason, to the fundamental PSO algorithm. Recent empirical studies have shown that selecting PSO control coefficients that are both order-$1$ and order-$2$ stable are vital to the performance of PSO \cite{ChrisClegSwarmStability}, and as such being able to easily obtain stability criteria for a PSO variant is an important issue for the field. 

The PSO variants this paper considers are those that can be rewritten as a finite sum of stochastically weighted difference vectors between a particle's position and swarm informers. Many PSO variants can be written in this stated form. The canonical PSO is in this form naturally, with  two particle informers, namely, the personal best position and the neighbourhood best position (or global best in the case of a fully connected swarm). The classic PSO variants, unified PSO (UPSO) \cite{UPSO} and fully informed PSO (FIPS)\cite{FIPS}, both use multiple informers, and can be written as a finite sum of stochastically weighted difference vectors. There is also a more recent trend of adding a third informer to PSO's update equation to guide a particle's movement based on information external to the swarm itself. Specifically, in the work of Scheepers \cite{Scheepers2018}, a variant of PSO for multi-objective optimization utilizes a third informer from the pareto front archive. A similar idea was also present in the work of Meier and Kramer \cite{PSORR3I}, where gradient based information was used to construct a third informer to assist PSO in the training of recurrent neural networks. 

The theorem presented in this paper, for obtaining stability criteria, also removes a common restriction present in existing stability work on PSO. Specifically, previous PSO stability results have provided stability criteria under the restriction that social and cognitive control coefficients are equal \cite{Jiang,weakStagnation,Poli1,BonayidNoStag}; such restrictions are not present when using the provided theorem. An additional theorem is also provided for obtaining the fixed points for the expectations and variance of particle positions.

A brief description of PSO, and its general form, is given in Section \ref{PSO_background}, followed by a summary of existing relevant PSO theory in Section \ref{PSO_theory}.  The theoretical derivations of criteria for stability along with the limit points for particle positions are provided in Section \ref{N informtors}. Section \ref{app_section} demonstrates the use of the stability theorem by deriving the stability criteria for three PSO variants. Additionally, Section \ref{app_section} provides the first order-$1$ and order-$2$ stability criteria of CPSO and UPSO without restriction on the relationship between control coefficients. A summary of the paper's findings is presented in Section \ref{concluiosn}.
%
%
%
%

\section{Particle Swarm Optimization}\label{PSO_background}
Particle swarm optimization (PSO) was originally inspired by the complex movement of birds in a flock. The variant of PSO this section focuses on is the CPSO algorithm \cite{InertiaIntroduction}.

The CPSO algorithm is defined as follows:
Let $f:\mathbb{R}^d\to \mathbb{R}$ be the objective function that the CPSO algorithm aims to find an optimum for, where $d$ is the dimensionality of the objective function. For the sake of simplicity, a minimization problem is assumed from this point onwards. Specifically, an optimum $\boldsymbol{o}\in\mathbb{R}^d$ is defined such that, for all $\boldsymbol{x}\in \mathbb{R}^d$, $f(\boldsymbol{o})\le f(\boldsymbol{x})$. In this paper the analysis focus is on objective functions where the optima exist.
Let $\Omega \left(t\right)$ be a set of $N$ particles in $\mathbb{R}^d$ at a discrete time step $t$. Then $\Omega \left(t\right)$  is said to be the particle swarm at time $t$. The position $\boldsymbol{x}_{i}$ of particle $i$ is updated using  
\small\begin{align}\label{eq:simple pos update in R^n}
	\boldsymbol{x}_{i}\left(t+1\right)
	&=\boldsymbol{x}_{i}\left(t\right)+\boldsymbol{v}_{i}\left(t+1\right),
\end{align} \normalsize
where the velocity update, $\boldsymbol{v}_{i}\left(t+1\right)$, is defined as
\small\begin{align}\label{eq:simple vel update in R^n}
	\boldsymbol{v}_{i}\left(t+1\right)&=w\boldsymbol{v}_{i}\left(t\right)+c_{1}\boldsymbol{r}_{1}(t)\otimes(\boldsymbol{y}_{i}(t)-\boldsymbol{x}_{i}\left(t\right)) \nonumber\\ &+c_{2}\boldsymbol{r}_{2}(t)\otimes(\hat{\boldsymbol{y}}_{i}(t)-\boldsymbol{x}_{i}\left(t\right)),
\end{align}\normalsize
where ${r}_{1,k}(t),{r}_{2,k}(t)\sim\ U\left(0,1\right)$ for all $t$ and $1\le k \le d$. The operator $\otimes$ is used to indicate component-wise multiplication of two vectors. The position $\boldsymbol{y}_{i}(t)$ represents the ``best'' position that particle $i$ has visited, where ``best'' means the location where the particle had obtained the lowest objective function evaluation. The position $\hat{\boldsymbol{y}}_{i}(t)$ represents the ``best'' position that the particles in the neighbourhood of the $i$-th particle have visited. The coefficients $c_1$, $c_2$, and $w$ are the cognitive, social, and inertia weights, respectively. A full algorithm description is presented in Algorithm \ref{alg:my alg}.

\begin{algorithm}
	\caption{PSO algorithm}
	\label{alg:my alg}
	\begin{algorithmic}
		\State Create and initialize a swarm, $\Omega \left(0\right)$, of $N$ particles uniformly within a predefined hypercube of dimension $d$.
		\State Let $f$ be the objective function.
		\State Let $\boldsymbol{y}_{i}$ represent the personal best position of particle $i$, initialized to $\boldsymbol{x}_{i}(0)$.
		\State Let $\hat{\boldsymbol{y}}_{i}$ represent the neighbourhood best position of particle $i$, initialized to $\boldsymbol{x}_{i}(0)$.
		\State Initialize $\boldsymbol{v}_{i}(0)$ to $\boldsymbol{0}$.
		\State Let $t=0$
		\Repeat
		\ForAll {particles $i=1,\cdots,N$ }
		\If{$f(\boldsymbol{x}_{i})<f(\boldsymbol{y}_{i})$}
		\State $\boldsymbol{y}_{i}=\boldsymbol{x}_{i}$
		\EndIf
		
		\ForAll {particles $\MYI$ with particle $i$ in their neighbourhood}
		\If{$f(\boldsymbol{y}_{i})<f(\hat{\boldsymbol{y}}_{\MYI})$}
		\State $\hat{\boldsymbol{y}}_{\MYI}=\boldsymbol{y}_{i}$
		\EndIf
		\EndFor
		
		\EndFor
		\State $t=t+1$
		\ForAll {particles $i=1,\cdots,N$ }
		\State update the velocity of particle $i$ using equation (\ref{eq:simple vel update in R^n})
		\State update the position of particle $i$ using equation (\ref{eq:simple pos update in R^n})
		\EndFor
		\Until{stopping condition is met}
	\end{algorithmic}
\end{algorithm}	


There are numerous PSO variants that alter equation \ref{eq:simple vel update in R^n} of the CPSO algorithm. The focus of this paper is on PSO variants whose velocity update equation can be rewritten into the following form:
\small\begin{align}
	&\boldsymbol{v}_{i}\left(t+1\right)=\boldsymbol{\theta}_0\otimes\boldsymbol{v}_{i}\left(t\right) + \sum_{\iota=1}^{I}\boldsymbol{\theta}_{\iota}\otimes\left(\boldsymbol{\zeta}_{\iota}\left(t\right)-\boldsymbol{x}_{i}\left(t\right)\right) \label{N_inf_V}\\
	&\boldsymbol{x}_{i}\left(t+1\right)=\boldsymbol{x}_{i}\left(t\right)+\boldsymbol{v}_{i}\left(t+1\right)\label{N_inf_X}
\end{align}\normalsize
where $\theta_{\iota,k}$ are arbitrary independent distributions with well defined mean and variance for each $0\le\iota\le I$, and $\boldsymbol{\zeta_{\iota}}$ represents each of the $I$ particle informers. In order to make referring to this general PSO formulation easier it is refereed to as N-Informer PSO (NIPSO).

\section{Current PSO Stability Analysis}\label{PSO_theory}

Almost all existing work has derived stability criteria directly for specific PSO variants. The CPSO algorithm has undergone the most theoretical stability analysis, from the earlier deterministic model works of \cite{KennedyContriction,TreleaTheoryPaper,VDBergTHeoryPaper} to the more recent stochastic works of \cite{Poli2,weakStagnation,BonayidNoStag,Order3,gerwien2020convergence}. A number of PSO variants have been directly studied \cite{Fernadez_CPSO_Arb_Dis,PSORR3I,ChrisCleghFIPS,ChrisCleghUPSO}. Recently, Cleghorn and Engelbrecht \cite{ChrisCleghPSOGeneralStabilityTheorem} proved \mbox{Theorem \ref{Main_contribution}} which allows for the derivation stability criteria for all PSO variants with the componentwise form:
\small\begin{align}\label{class_of_pso}
	& x_k(t+1)  =x_k(t)\alpha+x_k(t-1)\beta+\gamma_t
\end{align}\normalsize
where $\alpha$ and $\beta$ are well defined\footnote{In the context of this work a well defined random variable is one that has a mean and variance.} random variables, and $(\gamma_t)$ is a sequence of {well defined} random variables. The index $k$ indicates the vector component. The full theorem is now stated to assist in the subsequent derivations in Section \ref{N informtors}:
\begin{theorem}\label{Main_contribution}
	The following properties hold for all PSO variants of the form described in equation (\ref{class_of_pso}), where $E[\cdot]$ and $V[\cdot]$ are the expectation and variance operator respectively, and $\rho(\cdot)$ is the spectral radius of a matrix.
	\begin{enumerate}
		\item Assuming $\boldsymbol{i}_{t}$ converges, particle positions are order-$1$ stable for every initial condition if and only if $\rho(A)<1$, where
		\small\begin{align}\label{O1Matrix}
			A=\begin{bmatrix}E[\alpha] & E[\beta]  \\1& 0 \end{bmatrix}
			\text{ and } 	
			\boldsymbol{i}_t=\begin{bmatrix}E[\gamma_t] \\0\end{bmatrix}
		\end{align}
		\item The particle positions are order-$2$ stable if $\rho(\boldsymbol{B})<1$ and $(\boldsymbol{j}_{t})$ converges, where
		
		\small\begin{align}\label{main_matrix}
			&	\boldsymbol{B}=\begin{bmatrix}
				E[\alpha] & E[\beta] & 0 & 0 & 0\\
				1 & 0 & 0 & 0 & 0\\
				0 & 0 & E[\alpha^2] & E[\beta^2] & 2E[\alpha\beta]\\
				0 & 0 & 1 & 0 & 0\\
				0 & 0 & E[\alpha] & 0 & E[\beta] \\
			\end{bmatrix} \nonumber 
		\end{align} \normalsize
		and
		\small\begin{align}
			\boldsymbol{j}_t=\begin{bmatrix}
				E[\gamma_t] \\
				0 \\
				E[\gamma^2_t] \\
				0 \\
				0 \\
			\end{bmatrix} 
		\end{align}\normalsize under the assumption that the limits of $(E[\gamma_t\alpha])$ and $(E[\gamma_t\beta])$ exist.
		\item Assuming that $x(t)$ is order-$1$ stable, then the following is a necessary condition for order-$2$ stability:
		\begin{align}\label{Blacwell_extention}
			1-E\left[\alpha\right]-E\left[\beta\right] &\neq 0  \\\label{Con1_2}
			1-E\left[\alpha^2\right]-E\left[\beta^2\right]-\left(\frac{2E\left[\alpha\beta\right]E\left[\alpha\right]}{1-E\left[\beta\right]}\right)&> 0 
		\end{align}\normalsize
		\item The convergence of $E[\gamma_t]$ is a necessary condition for order-$1$ stability, and the convergence of both $E[\gamma_t]$ and $E[\gamma^2_t]$ is a necessary condition for  order-$2$ stability.
	\end{enumerate}	
\end{theorem}

While the generality of Theorem \ref{Main_contribution} is useful, it can make it potentially challenging for practitioners to quickly obtain stability criteria for their custom PSO variant from the theorem without a considerable amount of calculation. An example of the rigorous use of Theorem  \ref{Main_contribution} can be found in \cite{MGPSOSTAbility}. 

In order to reduce the burden on practitioners to derive stability criteria, a specialization of Theorem \ref{Main_contribution} to the class of PSOs described by equations \eqref{N_inf_V} and \eqref{N_inf_X} is proposed. The intention of the specialization is to make obtaining stability criteria as easy as possible, while still maintaining a sufficient degree of generality to cater for a range of variations in the PSO update equation formulation.
The overarching goal of the specialization is to reduce the need to perform  full stability analysis for most simple variants of PSO. In particular if a practitioner wished to augment the PSO update equations, ideally they should be able to quickly determine what the stability criteria of their bespoke variant is. Knowledge of the stability criteria is of vital importance for parameter tuning as it has been demonstrated that the stability of PSO particles is vital to the performance of PSO \cite{ChrisClegSwarmStability}.

At present all existing second order stability criteria published for CPSO have restricted the relationship between control coefficients. Specifically, the coefficients of CPSO have been restricted such that $c_1$ and $c_2$ were assumed either equal \cite{Poli2} or to have equal means and variances \cite{BonayidNoStag}. The theorem proved in the next section removes any such restriction, and can therefore produce stability criteria for arbitrary coefficient relationships. 

\section{Specialization to N-Informers}\label{N informtors}
This section provides the derivation of order-$1$ and \mbox{order-$2$} stability criteria for the class of  PSO variants as defined in equations \eqref{N_inf_V} and \eqref{N_inf_X}, which are collectively referred to as NIPSO. Furthermore, the order-$1$ and order-$2$ fixed points are derived.  
\begin{theorem}\label{PaperContribution}
	The following properties hold for all NIPSO combinations, under the non-stagnate distribution assumption for each informer.\footnote{	
		\textbf{Non-stagnant distribution assumption}:\\
		Let $\boldsymbol{\xi}_i(t)$ be an informer. It is assumed that $\boldsymbol{\xi}_i\left(t\right)$ is a random variable sampled from a time dependent distribution, such that  $\boldsymbol{\xi}_i(t)$  has a well defined expectation and variance for each $t$ and that $\lim\limits_{t\to \infty}E[\boldsymbol{\xi}_{i}(t)]$ and $\lim\limits_{t\to \infty}V[\boldsymbol{\xi}_{i}(t)]$ exist. A detailed justification of this modelling choice is given by Cleghorn and Engelbrecht \cite{ChrisCleghPSOGeneralStabilityTheorem}.
	} 
	
	\begin{enumerate}
		\item Particle positions are order-$1$ stable for every initial condition if and only if
		\begin{align}\label{THM_NOrder1Critera}
			-1<E[\theta_0]<1 
		\end{align}	
		and
		\begin{align}\label{THM_NOrder1Critera2}
			0<\sum_{\iota=1}^{I}E[\theta_{\iota}]<2(E[\theta_0]+1)
		\end{align}
		\item Particle positions are order-$2$ stable for every initial condition if and only if \footnote{The sufficient condition is not theoretically derived, as the inequality problem becomes intractable. Rather it is supported by extensive experimental evidence in line with the approach used by Bonyadi and Michalewicz \cite{BonayidNoStag} and Cleghorn and Engelbrecht \cite{ChrisCleghPSOGeneralStabilityTheorem} up to $I=25$. }
		\begin{align}
			\label{THM_newEq1}
			-1<\frac{E[\theta_0]}{\sqrt{1-V[\theta_0]}}<1 
		\end{align}
		and 
		\begin{align}
			0<\psi<\frac{-2\left(E[\theta_0]^2+V[\theta_0]-1\right)}{1-E[\theta_0]+\frac{\phi(1+E[\theta_0])}{\psi^2}}\label{Thn_newEq2}
		\end{align}
		where $\phi=\sum_{\iota=1}^{I} V[\theta_{\iota}]$ and $\psi=\sum_{\iota=1}^{I}E[\theta_{\iota}]$.
	\end{enumerate}
\end{theorem}
\textit{Proof (1)}:
Let the non-stagnate distribution assumption hold for each of the $I$ informers \footnote{Strictly speaking, only a well defined expectation and limit point of the informer is needed to prove part 1.} . Rewriting equations \eqref{N_inf_V} and \eqref{N_inf_X} into the general form of equation \eqref{class_of_pso} leads to:
\begin{align*}
	\alpha & = (1+\theta_0)-\sum_{\iota=1}^{I}\theta_{\iota}  \\
	\beta & =-\theta_0   \\
	\gamma_t &= \sum_{\iota=1}^{I}\theta_{\iota}\boldsymbol{\zeta}_{\iota}\left(t\right)
\end{align*}

In order to utilize part (1) of Theorem \ref{Main_contribution} to obtain the order-$1$ stability criteria the matrix $\boldsymbol{A}$ and the vector $\boldsymbol{i}_t$ must be constructed, as defined in equation \eqref{O1Matrix}. The required expectations are calculated as follows:
\begin{align}
	&E[\alpha]=1+E[\theta_0] -\sum_{\iota=1}^{I}E[\theta_{\iota}]\\
	&E[\beta]= -E[\theta_0 ]\\
	&E[\gamma_t]=\sum_{\iota=1}^{I}E[\theta_{\iota}]E[\boldsymbol{\zeta}_{\iota}\left(t\right)]
\end{align}
which leads to 
\begin{align}
	A=\begin{bmatrix}1+E[\theta_0] -\sum_{\iota=1}^{I}E[\theta_{\iota}] &\quad -E[\theta_0 ]  \\1& 0 \end{bmatrix} 
\end{align}
and
\begin{align}
	\boldsymbol{i}_t=\begin{bmatrix}\sum_{\iota=1}^{I}E[\theta_{\iota}]E[\boldsymbol{\zeta}_{\iota}\left(t\right)]\\0\end{bmatrix}.
\end{align}
Since $E[\theta_{\iota}]$ is well defined for each $\iota$ and  $E[\zeta_{\iota}\left(t\right)]$ is well defined and convergent for each $\iota$, by the non-stagnate distribution assumption, it follows that  ${i}_{t,0}$ is convergent and therefore $\boldsymbol{i}_{t}$ is convergent. In order to find the criteria needed to satisfy the condition $\rho(\boldsymbol{A})<1$, the eigenvalues of $\boldsymbol{A}$ are required and are calculated to be: 
\begin{align}
	\lambda_1,\lambda_2=\frac{\eta \pm \sqrt{\eta^2-4E[\theta_0 ]}}{2}
\end{align}
where $\eta=1+E[\theta_0] -\sum_{\iota=1}^{I}E[\theta_{\iota}]$. After some simplification it is found that $\rho(\boldsymbol{A})<1$ holds if and only if
\begin{align}\label{NOrder1Critera}
	-1<E[\theta_0]<1 \;\; \text{and} \;\; 0<\sum_{\iota=1}^{I}E[\theta_{\iota}]<2(E[\theta_0]+1).
\end{align}
It follows from part (1) of Theorem \ref{Main_contribution} that NIPSO is order-1 stable if and only if the criteria of equations \eqref{NOrder1Critera} and \eqref{THM_NOrder1Critera2} hold. \qed 

\textit{Proof (2)}:
Let the non-stagnate distribution assumption hold for each of the $I$ informers. In order to obtain the necessary conditions for order-$2$ stability, part 3 of Theorem \ref{Main_contribution} is utilized. A number of expectations are required to construct the matrix $\boldsymbol{B}$ and the vector $\boldsymbol{j}_t$. Specifically,
$E[\alpha^2]$, $E[\beta^2]$, and $E[\alpha\beta]$ are required, and calculated as
\small\begin{align}
	E[\alpha^2]&=E\left[\left(1+\theta_0-\sum_{\iota=1}^{I}\theta_{\iota}\right)^2\right] \nonumber\\ 
	&=1+2E[\theta_0] -2\sum_{\iota=1}^{I}E[\theta_{\iota}]\nonumber\\ \label{alphasqr1p2}
	&-2E[\theta_0] \sum_{\iota=1}^{I}E[\theta_{\iota}]+E\left[\left(\sum_{\iota=1}^{I}\theta_{\iota}\right)^2\right],
\end{align}
\normalsize where\small
\begin{align}
	E\left[\left(\sum_{\iota=1}^{I}\theta_{\iota}\right)^2\right]&=V\left[\sum_{\iota=1}^{I}\theta_{\iota}\right]+\left(\sum_{\iota=1}^{I}E[\theta_{\iota}]\right)^2 \nonumber\\
	&=\sum_{\iota=1}^{I}V[\theta_{\iota}]+\sum_{i\neq j}cov\left(\theta_i,\theta_j\right)+\left(\sum_{\iota=1}^{I}E[\theta_{\iota}]\right)^2 \nonumber\\
	&=\sum_{\iota=1}^{I}V[\theta_{\iota}]+\left(\sum_{\iota=1}^{I}E[\theta_{\iota}]\right)^2,\label{EExapand}
\end{align}\normalsize
since each $\theta_i$ are independent. Substituting equation \eqref{EExapand} back into equation \eqref{alphasqr1p2} leads to,
\begin{align}
	E[\alpha^2]&=1+2E[\theta_0] -2(1+[\theta_0])\sum_{\iota=1}^{I}E[\theta_{\iota}]\nonumber\\
	&+\sum_{\iota=1}^{I} V[\theta_{\iota}]+ \left(\sum_{\iota=1}^{I}E[\theta_{\iota}]\right)^2. \label{alphasqr2}
\end{align}
The expectation of $\beta^2$ and $\alpha\beta$ are easily calculated as:
\begin{align}
	E[\beta^2]&= E[\theta_0^2] =V[\theta_0]+E[\theta_0]^2\\
	E[\alpha\beta]&=E\left[-\theta_0 \left((1+\theta_0)-\sum_{\iota=1}^{I}\theta_{\iota}\right)\right] \nonumber\\
	&=-E[\theta_0]-V[\theta_0]-E[\theta_0]^2-E[\theta_0] \sum_{\iota=1}^{I}E[\theta_{\iota}].
\end{align}
For equation \eqref{Blacwell_extention}, in part 3 of Theorem \ref{Main_contribution}, to be satisfied the following condition must hold: 
\begin{align}\label{simpleCon}
	\psi=\sum_{\iota=1}^{I}E[\theta_{\iota}]\neq 0
\end{align} 
For equation \eqref{Con1_2}, in part 3 of Theorem \ref{Main_contribution}, to be satisfied the following condition must hold: 
\small
\begin{align*}
1+2E[\theta_0] +2(1+E[\theta_0])\psi-\phi- \psi^2
-V[\theta_0]+E[\theta_0]^2- \nonumber \\
\left( \frac{2\left(-E[\theta_0]-V[\theta_0]-E[\theta_0]^2-E[\theta_0] \psi\right)\left(1+E[\theta_0] -\psi\right)}{1+E[\theta_0 ]}  \right) >0,
\end{align*}
\normalsize
\noindent which is simplified using a method similar to that of Bonyadi and Michalewicz \cite{BonayidNoStag}, to equal the criteria of equations \eqref{THM_newEq1} and \eqref{Thn_newEq2}. The necessary condition of part $2$ of Theorem \ref{PaperContribution} is therefore proved. 

All that remains is to prove that satisfying the criteria of equations \eqref{THM_newEq1} and \eqref{Thn_newEq2} is in fact sufficient, and not only necessary, for order-$2$ stability. This is achieved by verifying that if the criteria of equations \eqref{THM_newEq1} and \eqref{Thn_newEq2} are satisfied then $\rho(\mathbf{B})<1$, from Theorem \ref{Main_contribution} part 2. All the expectations needed to construct matrix $\mathbf{B}$ have already been obtained while deriving the necessary condition.  In order to verify that $\rho(\mathbf{B})<1$ the empirical approach of Bonyadi and Michalewicz \cite{BonayidNoStag} and Cleghorn and Engelbrecht \cite{ChrisCleghPSOGeneralStabilityTheorem} is used. Specifically, for $I=1,2,\cdots,50$ informers the experimental procedure followed is: $I\times10^8$ random configurations representing $\{E[\theta_0],V[\theta_0],\cdots,E[\theta_I],V[\theta_I] \}$ are generated such that equations \eqref{THM_newEq1} and \eqref{Thn_newEq2} are satisfied. In all of the cases it was found that if equations \eqref{THM_newEq1} and \eqref{Thn_newEq2} were satisfied, then the condition $\rho(\mathbf{B})<1$ held. This finding is strong evidence that the criteria is sufficient for order-$2$ stability. \footnote{While the experimental verification was only done up to $50$ informers, there is no clear reason why it would fail to hold for higher informer counts. Practically speaking, a variant with more than $50$ informers seems unlikely.} \qed

\begin{theorem}
	The following properties hold for all NIPSO combinations:	
	\begin{enumerate}
		\item Under order-$1$ stability the fixed points of the particle position expectations are:
		\small\begin{align}
			{E}_{x_{i,k}}=\frac{\sum_{\iota=1}^{I}E\left[\theta_{\iota}\right]E\left[{{\zeta}}_{\iota,k}\right]}{\sum_{\iota=1}^{I}E\left[\theta_{\iota}\right]} \label{E_Fixed}
		\end{align}\normalsize
		where $E[{{\zeta}}_{\iota,k}]$ is the limit of $E[{\zeta}_{\iota,k}\left(t\right)]$.
		\item Under order-$1$ and order-$2$ stability, the fixed points of the particle position variances are:
		\small\begin{align}\label{var_fixed_point}
			&V_{x_{i,k}}=\\
			&\frac{\left(1+E[\theta_{0}])\right)\left(\kappa_1-2\kappa_2{E}_{x_{i,k}}+\kappa_3{E}^2_{x_{i,k}}\right)}{2\psi\left(1-E^2[\theta_{0}]-V[\theta_{0}]\right)-\phi\left(1+E[\theta_{0}]\right)+\psi^2\left(E[\theta_{0}]-1 \right)}\nonumber
		\end{align}\normalsize
		where
		\small\begin{align*}
			\kappa_1&=\sum_{\iota=1}^{I}\left( E^2[\theta_{\iota}]V[{\zeta}_{\iota,k}] +E^2[{\zeta}_{\iota,k}]V[\theta_{\iota}]+ V[\theta_{\iota}]V[{\zeta}_{\iota,k}]       \right),\\
			\kappa_2&=\sum_{\iota=1}^{I}V[\theta_{\iota}]E[{\zeta}_{\iota,k}],\quad
			\phi=\sum_{\iota=1}^{I} V[\theta_{\iota}],\quad
			\psi=\sum_{\iota=1}^{I}E[\theta_{\iota}],
		\end{align*}\normalsize
		with $E[{\zeta}_{\iota,k}]$ and $V[{\zeta}_{\iota,k}]$ as the the limit of $E[{\zeta}_{\iota,k}\left(t\right)]$ and $V[{\zeta}_{\iota,k}\left(t\right)]$ respectively.
		
	\end{enumerate}
\end{theorem}
\textit{Proof (1)}:
Under the assumption of order-$1$ stability each particle $i$ converges to a fixed point in expectation. Let such a fixed point be called $\boldsymbol{E}_{x_i}$. The fixed point is calculated by rewriting equations \eqref{N_inf_V} and \eqref{N_inf_X} into the following component-wise second order recurrence relation form:
\small\begin{align}
	{x}_{i,k}\left(t+1\right)=&
	{x}_{i,k}\left(t\right)(1+\theta_0)-\theta_0 x_{i,k}(t-1)\nonumber\\&+ \sum_{\iota=1}^{I}\theta_{\iota}\left({\zeta}_{\iota,k}\left(t\right)-{x}_{i,k}\left(t\right)\right).\label{N_inf_Recurance} 
\end{align}\normalsize
Applying the expectation operator leads to
\small\begin{align}
	E[{x}_{i,k}\left(t+1\right)]&=
	E[{x}_{i,k}\left(t\right)](1+E[\theta_0])-E[\theta_0]E[x_{i,k}(t-1)]\nonumber \\
	&+\sum_{\iota=1}^{I}E[\theta_{\iota}]\left(E[{\zeta}_{\iota,k}\left(t\right)]-E[{x}_{i,k}\left(t\right)]\right).\label{N_inf_RecuranceE} 
\end{align}\normalsize
Then by setting $E[x_{i,k}(t-1)]=E[x_{i,k}(t)]=E[x_{i,k}(t+1)]={E}_{x_{i,j}}$ and $E[{\zeta}_{\iota,k}\left(t\right)]$ to its limits $E[{\zeta}_{\iota,k}] $, equation \eqref{N_inf_RecuranceE} can be rearranged to find an explicit expression for ${E}_{x_{i,j}}$, thus obtaining equation \eqref{E_Fixed}.
\qed

\textit{Proof (2)}: Under the assumption of order-$1$ and order-$2$ stability each particle $i$ converges to a fixed point for each of the following sequences: $E[\boldsymbol{x}_{i}(t)]$, $E[\boldsymbol{x}_{i}(t)\boldsymbol{x}_{i}(t-1)]$, and $E[\boldsymbol{x}^2_{i}(t)]$. Let such fixed points be called $\boldsymbol{E}_{x_i}$, $\boldsymbol{E}_{x_i x_i}$, and $\boldsymbol{E}_{x^2_i}$ respectively as we will be working in the limit. First define 
\small\begin{align}\label{delta}
	\partial x_{i,k}(t)=x_{i,k}(t)-E[x_{i,k}(t)]=x_{i,k}(t)-{E}_{x_{i,k}}.
\end{align}\normalsize
It follows that $V[x_{i,k}(t)]=E[\partial^2 x_{i,k}(t)]$, where $\partial^2 x_{i,k}(t)$ denotes the square of equation \eqref{delta}. In order to obtain $V[x_{i,k}(t)]$, consider the class of recurrence relations as defined in equation \eqref{class_of_pso}, and that $\partial x_{i,k}(t)$ can be rewritten as
\small\begin{align}\label{partialE}
	\partial x_{i,k}(t)&= \alpha\partial x_{i,k}(t-1)+ \beta \partial x_{i,k}(t-2)+d_{i,k}(t-1)\\
	d_{i,k}(t-1)&=\gamma_{t-1}-{E}_{x_{i,k}}(1-\alpha-\beta).
\end{align}\normalsize
Squaring and applying the expectation operator to equation \eqref{partialE} leads to
\small\begin{align}\label{largeVequation}
	&E[\partial^2 x_{i,k}(t)] \nonumber\\
	& =E[\alpha^2]E[\partial^2 x_{i,k}(t-1)]+2E[\alpha\beta] E[\partial x_{i,k}(t-1)\partial x_{i,k}(t-2)]\nonumber \\
	&-E[d_{i,k}(t-1)]\left(2E[\alpha]E[\partial x_{i,k}(t-1)]+2E[\beta]E[\partial x_{i,k}(t-2)] \right)\nonumber \\
	&+E[\beta^2]E[\partial^2 x_{i,k}(t-2)]+E[d_{i,k}(t-1)]^2.
\end{align}\normalsize
In order to simplify equation \eqref{largeVequation} consider that
\small\begin{align}\label{helper1}
	E[\partial x_{i,k}(t)\partial x_{i,k}(t-1)]=E[\partial x_{i,k}(t-1)\partial x_{i,k}(t-2)]
	\end{align}\normalsize
	and
\small\begin{align}	
	\label{helper2}
	E[\partial x_{i,k}(t-2)]=E[\partial x_{i,k}(t-1)]=E[\partial{E}_{x_{i,k}}]=0.
\end{align}\normalsize
Now 
\small\begin{align} 
	&E[\partial x_{i,k}(t)\partial x_{i,k}(t-1)]\nonumber\\
	&=E[\alpha]E[\partial^2 x_{i,k}(t-1)]+E[\beta]E[\partial x_{i,k}(t-2)\partial x_{i,k}(t-1)]\nonumber\\
	&+E[d_{i,k}(t-1)]E[\partial x_{i,k}(t-1)]. \label{stepmultiple}
\end{align}\normalsize
Using equations \eqref{helper1} and \eqref{helper2}, equation \eqref{stepmultiple} can be rearranged to yield,
\small\begin{align} \label{stepmultipleComplete}
	E[\partial x_{i,k}(t)\partial x_{i,k}(t-1)]&=\frac{E[\alpha]E[\partial^2 x_{i,k}(t-1)]}{1-E[\beta]}.
\end{align}\normalsize
Now all that remains is to substitute equation \eqref{stepmultipleComplete} into equation \eqref{largeVequation}, and utilize the fact that $E[\partial^2 x_{i,k}(t-1)]=E[\partial^2 x_{i,k}(t)]$ (once again this is permissible in the limit), to obtain
\small\begin{align}\label{generalVaraince}
	E[\partial^2 x_{i,k}(t)]=\frac{E[d_{i,k}(t-1)^2]}{1-E[\alpha^2]-E[b^2]-\dfrac{2E[\alpha\beta][\alpha]}{1-E[\beta]}}.
\end{align}\normalsize
Equation \eqref{generalVaraince} represents the variance fixed point for the large class of PSOs. However, our focus is on the case where 
\small\begin{align*}
	\alpha & = (1+\theta_0)-\sum_{\iota=1}^{I}\theta_{\iota}  \\
	\beta & =-\theta_0   \\
	\gamma_t &= \sum_{\iota=1}^{I}\theta_{\iota}{\zeta}_{\iota,k}\left(t\right).
\end{align*}\normalsize
Substituting these specific $\alpha$, $\beta$, and $\gamma_t$ into equation \eqref{generalVaraince} and performing a substantial amount of simplification (which is omitted for the sake of brevity), leads to equation \eqref{var_fixed_point} as was required to be proved. \qed

\section{Application of Stability Results}\label{app_section}
In this section a number of existing stability criteria are re-derived to demonstrate how Theorem \ref{PaperContribution} can be easily applied to rapidly obtain stability criteria.  Furthermore, previous stability results have considerably restricted the allowable relationship between control coefficients, for example $c_1=c_2$, this limitation is present in this section. All derived criteria contained in the section have no restriction on the coefficient relations, and as such are a novel contribution in addition to illustrating the ease of using Theorem \ref{PaperContribution}.

Stability criteria for CPSO, fully informed PSO, and unified PSO are derived in sections \ref{IPSO_S}, \ref{FIPSO_S}, and \ref{UPSO_S} respectively. 

\subsection{Canonical PSO} \label{IPSO_S}
\begin{figure}[!ht] 
	\subfloat[Order-2 stable regions for $w=0.4$ to $0.9$.]{
		\includegraphics[width=0.485\textwidth]{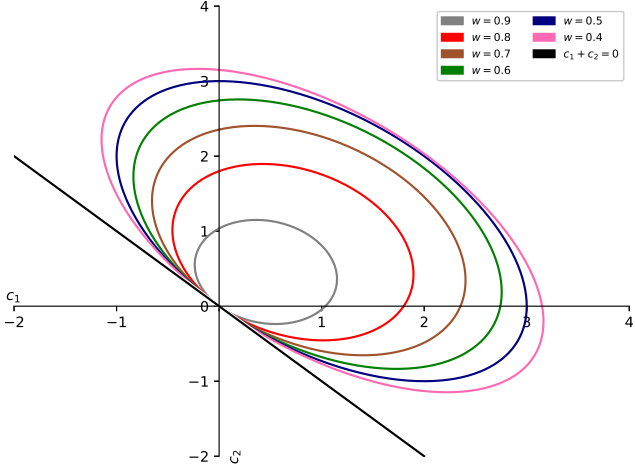}%
	}
	\hfill
	\subfloat[Order-2 stable regions for $w=-0.9$ to $0.3$.]{
		\includegraphics[width=0.485\textwidth]{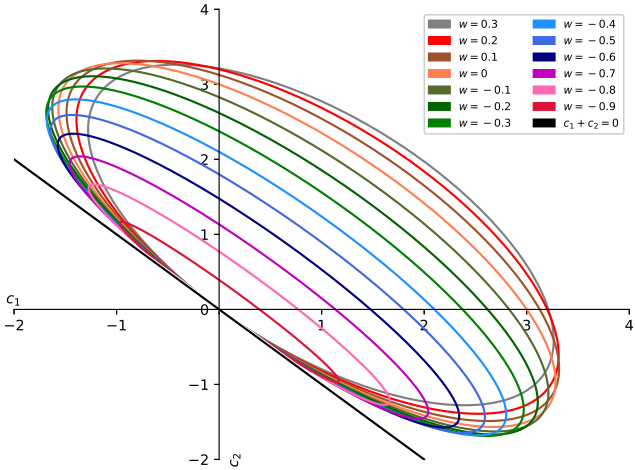}%
	}
	\caption{Order-2 stable regions of CPSO under fixed inertia values. The interior region of the elliptic shapes correspond to where equation \eqref{CPSO} is satisfied for a given $w$.}
	\label{CPSO_regions}
\end{figure}

Consider the case of the CPSO algorithm, as defined by equations \eqref{eq:simple pos update in R^n} and \eqref{eq:simple vel update in R^n}. After dropping the particle and component indices, without loss of generality, the stability criteria for CPSO can be obtained by using two informers with $\theta_0=w$, $\theta_1=c_1r_1$, and $\theta_2=c_2r_2$. 
In order to utilize Theorem \ref{PaperContribution}, $\psi$ and $\phi$ are required and calculated as: 

\small\begin{align}
	\psi=\sum_{\iota=1}^{2}E[\theta_{\iota}]=\frac{c_1}{2}+\frac{c_2}{2}, \quad 
	\phi=\sum_{\iota=1}^{2} V[\theta_{\iota}]=\frac{c^2_1}{12}+\frac{c^2_2}{12}.
\end{align}\normalsize
Substituting $\psi$ and $\phi$ into the criteria of equations \eqref{THM_NOrder1Critera}, \eqref{THM_NOrder1Critera2}, \eqref{THM_newEq1}, and \eqref{Thn_newEq2} the following criteria for order-$1$ and order-$2$ stability are obtained:
\small\begin{align}\label{CPSO}
	-1<w<1 \quad \text{and}\quad
	0<c_1+c_2<\frac{4\left(1-w^2\right)}{1-w+\frac{\left(c^2_1+c^2_2\right)(1+w)}{3(c_1+c_2)^2}}.
\end{align}\normalsize
The criteria in equation \eqref{CPSO} is the first time CPSO's full order-$1$ and order-$2$ stability criteria has not been simplified to the case where $c_1=c_2$. If the simplification is reimposed, the following commonly reported form reappears:
\begin{align}\label{CPSOS}
	-1<w<1 \quad \text{and}\quad
	0<c_1+c_2<\frac{24\left(1-w^2\right)}{7-5w}.
\end{align}\normalsize
It is interesting to observe that the weighting between $c_1$ and $c_2$ has a direct influence of the size and shape of the stability region as illustrated in figure \ref{CPSO_regions}, where the cross-sections of the stability region, with fixed inertia values, are shown. Additionally, Figure \ref{CPSO_regions} demonstrates that using equation \eqref{CPSOS} without the knowledge of the restrictions of $c_1=c_2$ can lead to the misclassification of stable parameter configurations. 

\subsection{Fully Informed PSO } \label{FIPSO_S}
The FIPS algorithm is an early PSO variant proposed by Kennedy and Mendes \cite{FIPS}, based on the observation that in human society individuals are not influenced by only a single individual, but rather by a statistical
summary of the state of their neighbourhood. In the FIPS algorithm, the velocity equation of CPSO is altered such that each particle is influenced by all its neighbours. Specifically, the velocity update equation for FIPS is:
\small\begin{align}\label{FIPS update}
	\boldsymbol{v}_{i}\left(t+1\right) &= w\boldsymbol{v}_{i}\left(t\right)+ \sum_{m=1}^{|\mathcal{N}_i|} \boldsymbol{\gamma}_m\otimes\frac{(\boldsymbol{y}_{m}(t)-\boldsymbol{x}_{i}\left(t\right))}{|\mathcal{N}_i|},
\end{align}\normalsize
where $\mathcal{N}_i$ is the set of particles in particle $i$'s neighbourhood, $\boldsymbol{y}_m(t)\in \mathcal{N}_i$, and ${\gamma}_{m,k}\sim U\left(0,\hat{c}\right)$, were $c_1+c_2=\hat{c}$.

After dropping the particle and component indices, without loss of generality, the stability criteria for FIPS can be obtained by considering $I=|\mathcal{N}|$ informers and setting $\theta_0=w$ and $\theta_{\iota}=\frac{{\gamma}_\iota}{|\mathcal{N}|}$ for $1\le \iota\le |\mathcal{N}|$.  The following calculations are required to use Theorem \ref{PaperContribution}:


\small\begin{align}
	\psi=\sum_{\iota=1}^{I}E[\theta_{\iota}]=\sum_{\iota=1}^{|\mathcal{N}|} \frac{E[{\gamma}_\iota]}{|\mathcal{N}|}=\sum_{\iota=1}^{|\mathcal{N}|} \frac{\hat{c}}{2|\mathcal{N}|}=\frac{\hat{c}}{2}
\end{align}\normalsize
and
\small\begin{align}
	\phi=\sum_{\iota=1}^{I} V[{\gamma}_{\iota}]=\sum_{\iota=1}^{|\mathcal{N}|} V\left[\frac{{\gamma}_\iota}{ |\mathcal{N}|}\right]=\sum_{\iota=1}^{|\mathcal{N}|} \frac{\hat{c}^2}{12|\mathcal{N}|^2}=\frac{\hat{c}^2}{12|\mathcal{N}|}.
\end{align}\normalsize
Substituting $\psi$ and $\phi$ into the criteria of equations \eqref{THM_NOrder1Critera}, \eqref{THM_NOrder1Critera2}, \eqref{THM_newEq1}, and \eqref{Thn_newEq2} the following criteria for order-$1$ and order-$2$ stability are obtained:


\small\begin{align}\label{FIPSO}
	-1<w<1 \quad \text{and}\quad
	0<\frac{\hat{c}}{2}<\frac{12\left(1-w^2\right)}{3|\mathcal{N}|+1+w(1-3|\mathcal{N}|)}.
\end{align}\normalsize
The derived criteria is in agreement with existing criteria of both Cleghorn and
Engelbrecht \cite{ChrisCleghFIPS} and  Garc\'{i}a-Gonzalo and Fern\'{a}ndez-Martinez \cite{FernandezGoodpaper}, but are obtained with minimal calculations, and under a weaker modelling assumption.

\subsection{Unified PSO} \label{UPSO_S}
The UPSO algorithm was designed by Parsopoulos and Vrahatis \cite{UPSO} as a weighted merger between the local best PSO and the global best PSO. The PSO variants utilizes the additional control parameter, \mbox{$u\in[0,1]$}, called the unification factor, to control the importance placed on either the global best PSO update or the local best PSO. Specifically, the update equation for UPSO are:
\small\begin{align}
	\boldsymbol{g}_i\left(t+1\right)&= w \boldsymbol{v}_i\left(t\right)+c_1 \boldsymbol{r_1}\otimes(\boldsymbol{y}_i(t)-\boldsymbol{x}_i\left(t\right))\nonumber\\ &+c_2 \boldsymbol{r_2}\otimes(\mathbf{g}(t)-\boldsymbol{x}_i\left(t\right)) \label{UPSOVupdate_g}\\
	\boldsymbol{l}_i\left(t+1\right)&= w \boldsymbol{v}_i\left(t\right)+c_1 \boldsymbol{r'_1}\otimes(\boldsymbol{y}_i(t)-\boldsymbol{x}_i\left(t\right))\nonumber\\&+ c_2 \boldsymbol{r'_2}\otimes(\hat{\boldsymbol{y}}_i(t)-\boldsymbol{x}_i\left(t\right)) \label{UPSOVupdate_l}\\
	\boldsymbol{v}_i\left(t+1\right)&= u\boldsymbol{g}_i(t+1)+\left(1-u\right)\boldsymbol{l}_i(t+1) \label{UPSOVupdate}\\
	\boldsymbol{x}_i\left(t+1\right)&= \boldsymbol{x}_i\left(t\right)+\boldsymbol{v}_i\left(t+1\right)\label{UPSOPupdate}, 
\end{align}\normalsize
where $r_{1,k},r_{2,k},r'_{1,k}, r'_{2,k} \sim U\left(0,1\right)$, and both $\mathbf{y}_i(t)$ and $\hat{\mathbf{y}}_i(t)$ are defined as before with the addition of $\mathbf{g}(t)$ as the global best position with the swarm at time step $t$.

\begin{figure}[!ht]
	\subfloat[Order-2 stable regions for $u=0.25$  and $w=0.4$ to $0.9$.]{%
		\includegraphics[width=0.485\textwidth]{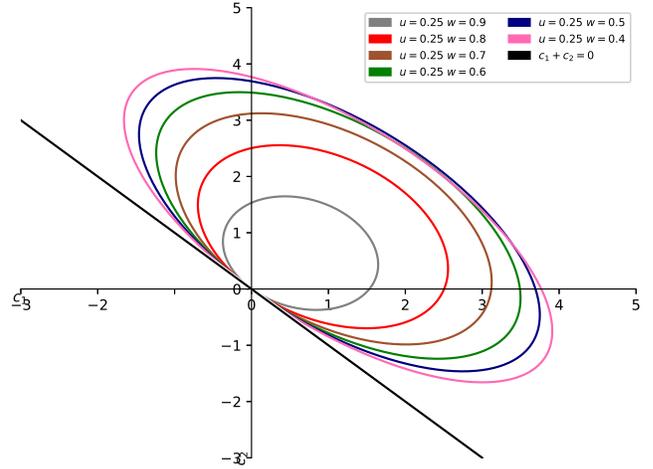}
	}
	\hfill
	\subfloat[Order-2 stable regions for $u=0.25$ and $w=-0.9$ to $0.3$.]{%
		\includegraphics[width=0.485\textwidth]{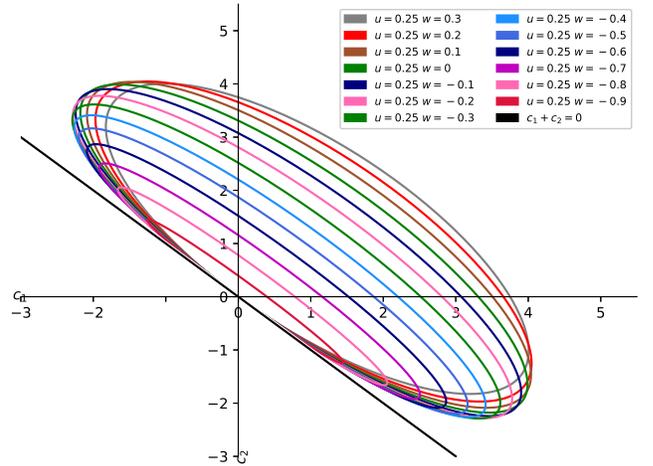}
	}
	\caption{Order-2 stable regions of UPSO under fixed inertia and unification values. The interior region of the elliptic shapes correspond to where equations \eqref{order2UPSO} and \eqref{order2UPSO2} are satisfied for a given $w$ and $u$.}
	\label{UPSO_regions}
\end{figure}

Without loss of generality the particle and component wise index are dropped again. In order to rewrite UPSO into the NIPSO form, substitute equations \eqref{UPSOVupdate_g} and \eqref{UPSOVupdate_l} into the velocity update equation \eqref{UPSOVupdate} to arrive at
\small\begin{align}
	v\left(t+1\right)&=w v\left(t\right)  
	+ c_1(ur_1+(1-u){r'_1})(y(t)-x\left(t\right) ) \label{UPSO_sub_velocity}   \\
	&+ c_2ur_2(g(t)-x\left(t\right) )   
	+ c_2(1-u)r'_2(\hat{y}_i(t)-x\left(t\right) ). \nonumber 
\end{align}\normalsize
Now equation \eqref{UPSO_sub_velocity} is in the NIPSO form with $I=3$ and  $\theta_0=w$, $\theta_{1}=c_1(ur_1+(1-u){r'_1})$, $\theta_{2}=c_2ur_2$, and $\theta_3=c_2(1-u)r'_2$. In order to calculate $\psi$ the following additional terms are required:
\small\begin{align}
	E[\theta_{1}]&=c_1uE[r_1]+c_1(1-u)E[r'_1]\nonumber\\
	&=\frac{c_1u}{2}+\frac{c_1(1-u)}{2} =\frac{c_1}{2}  \label{psi1}\\
	E[\theta_{2}]&=c_2uE[r_2]=\frac{c_2u}{2}  \label{psi2}\\
	E[\theta_{3}]&=c_2(1-u)E[r'_2]=\frac{c_2(1-u)}{2}  \label{psi3}.
\end{align}\normalsize
The summation of equations \eqref{psi1}, \eqref{psi2}, and \eqref{psi3} leads to
\begin{align}
	\psi&=\sum_{\iota=1}^{3}E[\theta_{\iota}]=\frac{c_1}{2} +\frac{c_2u}{2}+\frac{c_2(1-u)}{2}
	= \frac{c_1+c_2}{2} .
\end{align}
In order to calculate $\psi$ the following additional terms are required:
\small\begin{align}
	V[\theta_{1}]&=c_1^2V[ur_1+(1-u){r'_1}] \nonumber \\
	&=c_1^2\left(  u^2V[r_1]+(1-u)^2V[r_2]+2u(1-u)COV[r_1,r'_2]      \right) \nonumber \\
	&=c_1^2\left(  \frac{u^2}{12}+\frac{(1-u)^2}{12} \right)=c_1^2\left(\frac{u^2+(1-u)^2}{12}\right)\label{phi1}\\
	V[\theta_{2}]&=V[c_2ur_2]= \frac{c_2^2 u^2}{12} \label{phi2}\\
	V[\theta_{2}]&=V[c_2(1-u)r'_2]= \frac{c_2^2 (1-u)^2}{12}. \label{phi3}
\end{align}\normalsize
The summation of equations \eqref{phi1}, \eqref{phi2}, and \eqref{phi3} leads to 
\small\begin{align}
	\phi=\sum_{\iota=1}^{3} V[\theta_{\iota}]=\frac{(c_1^2+c_2^2)\left(u^2+(1-u)^2\right)}{12}.
\end{align}\normalsize
Substituting $\psi$ and $\phi$ into the criteria of equations \eqref{THM_NOrder1Critera}, \eqref{THM_newEq1}, and \eqref{Thn_newEq2} the following criteria for order-$1$ and order-$2$ stability are obtained:
\small\begin{align}\label{order2UPSO}
	-1<w<1 \\\label{order2UPSO2}
	0<c_1+c_2 <\frac{4\left(1-w^2\right)}{1-w+\frac{(c_1^2+c_2^2)\left(u^2+(1-u)^2\right)(1+w)}{3(c_1+c_2 )^2}}.
\end{align}\normalsize
The criteria of equations \eqref{order2UPSO} and \eqref{order2UPSO2} is the first derivation of full USPO stability criteria without artificial restrictions on the control coefficients. As with the CPSO case, in \mbox{Section \ref{IPSO_S}}, the weighting between $c_1$and $c_2$ has a clear influence on the size and shape of stability region, as illustrated in \mbox{Figure \ref{UPSO_regions}}, where the cross-sections of the stability region, with fixed inertia values is shown.

In the restricted case where $c_1=c_2$ is considered, the following criteria are obtained:
\small\begin{align}\label{order2UPSOE}
	-1<w<1 \\
	0<c_1+c_2 <\frac{24\left(1-w^2\right)}{7-5w+2(u^2-u)(1+w)}.
\end{align}\normalsize
which is in agreement with the derived criteria of Cleghorn and Engelbrecht \cite{ChrisCleghUPSO} with minimal calculations needed, and under a weaker modelling assumption.

\section{Conclusion} \label{concluiosn}	
This paper derives general theorems for rapidly obtaining order-$1$ and order-$2$ stability criteria and fixed points for a class of PSO variants. Specifically, PSO variants that can be rearranged into a sum of difference vectors between informers and the current particle positions, are catered for. From this general derivation, stability criteria can be obtained for a set of custom PSO variants in a direct manner without substantial mathematical calculation. Given the direct linkage between PSO performance and the satisfaction of order-$1$ and order-$2$ stability criteria, the theorems provided in this paper will be directly applicable to the PSO community as a whole.

Furthermore, the proved theorems allows for stability criteria to be derived without unnecessary restrictions on the relationship between control coefficients. In this vein, stability criteria for both the canonical PSO and the unified PSO are, for the first time, derived without restrictions on the relationship between control coefficients in this paper.  

\bibliographystyle{IEEEtran}
\bibliography{IEEEabrv,Bibliography}

\begin{thebibliography}{10}
\providecommand{\url}[1]{#1}
\csname url@samestyle\endcsname
\providecommand{\newblock}{\relax}
\providecommand{\bibinfo}[2]{#2}
\providecommand{\BIBentrySTDinterwordspacing}{\spaceskip=0pt\relax}
\providecommand{\BIBentryALTinterwordstretchfactor}{4}
\providecommand{\BIBentryALTinterwordspacing}{\spaceskip=\fontdimen2\font plus
\BIBentryALTinterwordstretchfactor\fontdimen3\font minus
  \fontdimen4\font\relax}
\providecommand{\BIBforeignlanguage}[2]{{%
\expandafter\ifx\csname l@#1\endcsname\relax
\typeout{** WARNING: IEEEtran.bst: No hyphenation pattern has been}%
\typeout{** loaded for the language `#1'. Using the pattern for}%
\typeout{** the default language instead.}%
\else
\language=\csname l@#1\endcsname
\fi
#2}}
\providecommand{\BIBdecl}{\relax}
\BIBdecl

\bibitem{OriginalPSOPaper}
J.~Kennedy and R.~Eberhart, ``Particle swarm optimization,'' in
  \emph{Proceedings of the IEEE International Joint Conference on Neural
  Networks}.\hskip 1em plus 0.5em minus 0.4em\relax Piscataway, NJ: IEEE Press,
  1995, pp. 1942--1948.

\bibitem{PoliAppReview}
R.~Poli, ``Analysis of the publications on the applications of particle swarm
  optimisation,'' \emph{Journal of Artificial Evolution and Applications}, vol.
  2008, pp. 1--10, 2008.

\bibitem{FirstTheoryPaper}
E.~Ozcan and C.~Mohan, ``Analysis of a simple particle swarm optimization
  system,'' \emph{Intelligent Engineering Systems through Artificial Neural
  Networks}, vol. volume 8, pp. 253--258, 1998.

\bibitem{KennedyContriction}
M.~Clerc and J.~Kennedy, ``The particle swarm{-}explosion, stability, and
  convergence in a multidimensional complex space,'' \emph{IEEE Transactions on
  Evolutionary Computation}, vol.~6, no.~1, pp. 58--73, 2002.

\bibitem{Jiang}
M.~Jiang, Y.~Luo, and S.~Yang, ``Stochastic convergence analysis and parameter
  selection of the standard particle swarm optimization algorithm,''
  \emph{Information Processing Letters}, vol. 102, no.~1, pp. 8--16, 2007.

\bibitem{Fernadez_CPSO_Arb_Dis}
E.~Garc{\'i}a-Gonzalo and J.~Fern{\'a}ndez-Martinez, ``Convergence and
  stochastic stability analysis of particle swarm optimization variants with
  generic parameter distributions,'' \emph{Applied Mathematics and
  Computation}, vol. 249, pp. 286--302, 2014.

\bibitem{ChrisClegh}
C.~Cleghorn and A.~Engelbrecht, ``A generalized theoretical deterministic
  particle swarm model,'' \emph{Swarm Intelligence}, vol.~8, no.~1, pp. 35--59,
  2014.

\bibitem{weakStagnation}
Q.~Liu, ``Order-2 stability analysis of particle swarm optimization,''
  \emph{Evolutionary Computation}, vol.~23, no.~2, pp. 187--216, 2015.

\bibitem{ChrisCleghPSOGeneralStabilityTheorem}
C.~W. Cleghorn and A.~P. Engelbrecht, ``Particle swarm stability: a theoretical
  extension using the non-stagnate distribution assumption,'' \emph{Swarm
  Intelligence}, vol.~12, no.~1, pp. 1--22, 2018.

\bibitem{CleghWhatIsStability}
C.~Cleghorn, ``Particle swarm optimization: Understanding order-2 stability
  guarantees,'' in \emph{Proceedings of the International Conference on the
  Applications of Evolutionary Computation}.\hskip 1em plus 0.5em minus
  0.4em\relax Switzerland: Springer, 2019, pp. 535--549.

\bibitem{InertiaIntroduction}
Y.~Shi and R.~Eberhart, ``A modified particle swarm optimizer,'' in
  \emph{Proceedings of the IEEE Congress on Evolutionary Computation}.\hskip
  1em plus 0.5em minus 0.4em\relax Piscataway, NJ: IEEE Press, 1998, pp.
  69--73.

\bibitem{ChrisClegSwarmStability}
C.~Cleghorn and A.~Engelbrecht, ``Particle swarm optimizer: The impact of
  unstable particles on performance,'' in \emph{Proceedings of the IEEE
  Symposium Series on Swarm Intelligence}.\hskip 1em plus 0.5em minus
  0.4em\relax Piscataway, NJ: IEEE Press, 2016, pp. 1--7.

\bibitem{UPSO}
K.~Parsopoulos and M.~Vrahatis, ``{UPSO}: A unified particle swarm optimization
  scheme,'' in \emph{Proceedings of the International Conference on
  Computational Methods in Sciences and Engineering}.\hskip 1em plus 0.5em
  minus 0.4em\relax Netherlands: VSP International Science Publishers, 2004,
  pp. 868--873.

\bibitem{FIPS}
J.~Kennedy and R.~Mendes, ``Neighborhood topologies in fully-informed and
  best-of-neighborhood particle swarms,'' in \emph{Proceedings of the IEEE
  International Workshop on Soft Computing in Industrial Applications}.\hskip
  1em plus 0.5em minus 0.4em\relax Piscataway, NJ: IEEE Press, 2003, pp.
  45--50.

\bibitem{Scheepers2018}
C.~Scheepers, ``Multi-guided particle swarm optimization: A multi-objective
  particle swarm optimizer,'' Doctor's dissertation, University of Pretoria,
  2018.

\bibitem{PSORR3I}
A.~Meier and O.~Kramer, ``Recurrent neural network-predictions for pso in
  dynamic optimization,'' in \emph{Proceedings of the Genetic and Evolutionary
  Computation Conference}.\hskip 1em plus 0.5em minus 0.4em\relax New York, NY:
  ACM Press, 2018, pp. 29--36.

\bibitem{Poli1}
R.~Poli and D.~Broomhead, ``Exact analysis of the sampling distribution for the
  canonical particle swarm optimiser and its convergence during stagnation,''
  in \emph{Proceedings of the Genetic and Evolutionary Computation
  Conference}.\hskip 1em plus 0.5em minus 0.4em\relax New York, NY: ACM Press,
  2007, pp. 134--141.

\bibitem{BonayidNoStag}
M.~Bonyadi and Z.~Michalewicz, ``Stability analysis of the particle swarm
  optimization without stagnation assumption,'' \emph{IEEE Transactions on
  Evolutionary Computation}, vol.~20, no.~5, pp. 814--819, 2016.

\bibitem{TreleaTheoryPaper}
I.~Trelea, ``The particle swarm optimization algorithm: Convergence analysis
  and parameter selection,'' \emph{Information Processing Letters}, vol.~85,
  no.~6, pp. 317--325, 2003.

\bibitem{VDBergTHeoryPaper}
F.~Van~den Bergh and A.~Engelbrecht, ``A study of particle swarm optimization
  particle trajectories,'' \emph{Information Sciences}, vol. 176, no.~8, pp.
  937--971, 2006.

\bibitem{Poli2}
R.~Poli, ``Mean and variance of the sampling distribution of particle swarm
  optimizers during stagnation,'' \emph{IEEE Transactions on Evolutionary
  Computation}, vol.~13, no.~4, pp. 712--721, 2009.

\bibitem{Order3}
W.~Dong and R.~Zhang, ``Order-3 stability analysis of particle swarm
  optimization,'' \emph{Information Sciences}, vol. 503, pp. 508--520, 2019.

\bibitem{gerwien2020convergence}
M.~Gerwien, R.~Vo{\ss}winkel, and H.~Richter, ``Convergence analysis of
  particle swarm optimization using stochastic lyapunov functions and
  quantifier elimination,'' \emph{arXiv preprint arXiv:2002.01673}, 2020.

\bibitem{ChrisCleghFIPS}
C.~Cleghorn and A.~Engelbrecht, ``Fully informed particle swarm optimizer:
  Convergence analysis,'' in \emph{Proceedings of the IEEE Congress on
  Evolutionary Computation}.\hskip 1em plus 0.5em minus 0.4em\relax Piscataway,
  NJ: IEEE Press, 2015, pp. 164--170.

\bibitem{ChrisCleghUPSO}
------, ``Unified particle swarm optimizer: Convergence analysis,'' in
  \emph{Proceedings of the IEEE Congress on Evolutionary Computation}.\hskip
  1em plus 0.5em minus 0.4em\relax Piscataway, NJ: IEEE Press, 2016, pp.
  448--454.

\bibitem{MGPSOSTAbility}
C.~Cleghorn, C.~Scheepers, and A.~Engelbrecht, ``Stability analysis of the
  multi-objective multi-guided particle swarm optimizer,'' in \emph{Proceedings
  of International Swarm Intelligence Conference (ANTS), Swarm
  Intelligence}.\hskip 1em plus 0.5em minus 0.4em\relax Switzerland: Springer
  International Publishing, 2018, pp. 201--212.

\bibitem{FernandezGoodpaper}
E.~Garc{\'i}a-Gonzalo and J.~Fern{\'a}ndez-Martinez, ``Convergence and
  stochastic stability analysis of particle swarm optimization variants with
  generic parameter distributions,'' \emph{Applied Mathematics and
  Computation}, vol. 249, pp. 286--302, 2014.

\end{thebibliography}

\balance



\end{document}